\documentclass[10pt,twocolumn,letterpaper]{article}

\usepackage{cvpr}              
\usepackage{multirow}
\usepackage{algorithm}
\usepackage[algo2e,ruled,linesnumbered]{algorithm2e}
\definecolor{cvprblue}{rgb}{0.21,0.49,0.74}
\usepackage[pagebackref,breaklinks,colorlinks,allcolors=cvprblue]{hyperref}

\def\paperID{} 
\def\confName{CVPR}
\def\confYear{2026}

\title{DreamSR: Towards Ultra-High-Resolution Image Super-Resolution via a Receptive-Field Enhanced Diffusion Transformer}

\author{Qingji Dong$^{1}$\quad Hang Dong$^{1}$\thanks{Corresponding Author.}\quad Mingqin Chen$^{1}$\quad Rui Zhang$^{1}$\quad Yitong Wang$^{1}$ \\
	$^{1}$ ByteDance Inc. \\
}

\begin{document}
\maketitle
\begin{figure*}[!htbp]
    \centering
    \includegraphics[width=\linewidth]{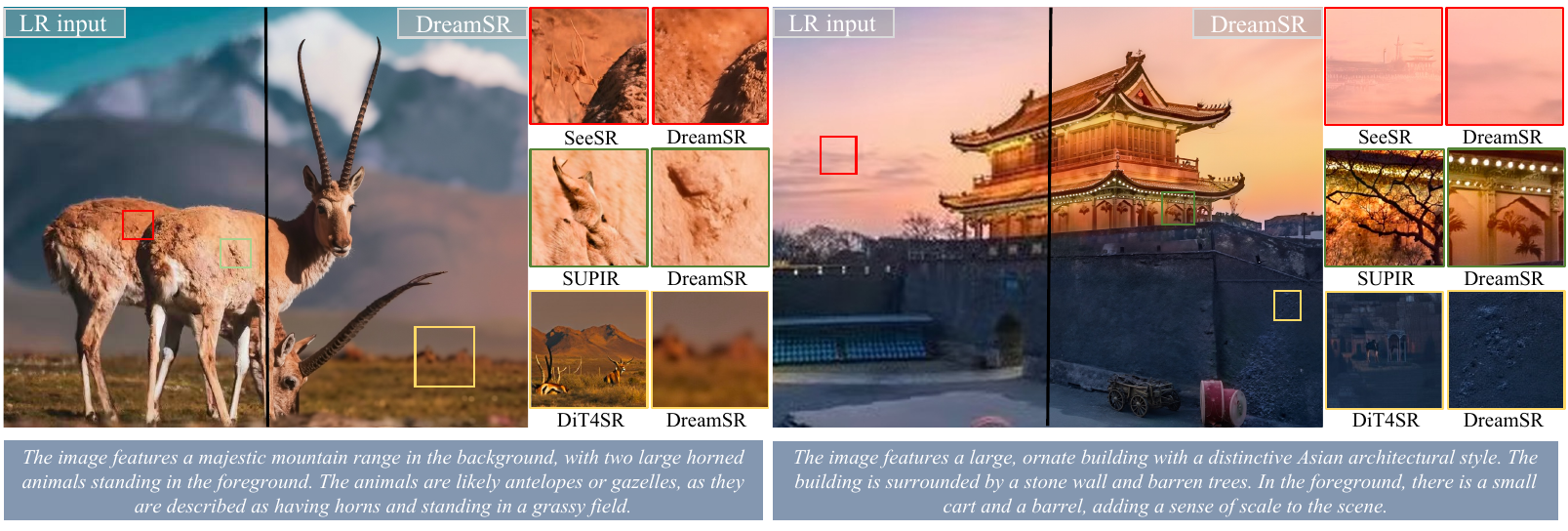}
    \vspace{-6mm}
    \caption{Example of local over-generation in patch-wise inference for high-resolution images. When existing methods adopt patch-wise inference strategy, inconsistent semantic details may appear, due to the misalignment between the global prompt and local patches. On the other hand, the proposed DreamSR can effectively suppresses such artifacts and restores fine-grained textures with high fidelity.}
    \vspace{-2mm}
    \label{showcase}
\end{figure*}

\begin{abstract}
Large-scale pre-trained diffusion models have been extensively adopted for real-world image Super-Resolution because of their powerful generative priors {through textual guidance}. However, when super-resolving high-resolution images with patch-wise inference strategy, most existing diffusion-based SR methods tend to suffer from over-generation, {due to the misalignment between the global prompt from LR image and the incomplete semantic information of local patches during each inference step.} On the other hand, most existing methods also failed to generate detailed texture in local patches due to the overemphasis on global generation capabilities in network designs and training strategies.
To address this issue, we present DreamSR, a novel SR model that suppresses local over-generation and improves fine-detail synthesis, thereby achieving visually faithful results with ultra-high-quality details. 
Specifically, we propose a dual-branch MM-ControlNet, where the ControlNet generates local textual feature with patch-level prompts while the pre-trained DiT provides global textual feature with global prompts, thereby mitigating over-generation and ensuring semantic consistency across patches.
We also design a comprehensive training strategy with stage-specific data processing pipelines and a Receptive-Field Enhancement strategy, enhancing the model’s capability to capture patch information and effectively restore local textures.
Extensive experiments demonstrate that DreamSR outperforms state-of-the-art methods, providing high-quality SR results. Code and model are available at \href{https://github.com/jerrydong0219/DreamSR}{https://github.com/jerrydong0219/DreamSR}.
\end{abstract}    
\section{Introduction}
\label{sec:intro}

{Real-world Image Super-Resolution (Real-ISR) aims to reconstruct high-resolution (HR) images from low-resolution (LR) inputs with unknown degradation. This task requires the model not only to effectively remove diverse degradations but also to generate perceptually realistic details, especially in challenging ultra-high-resolution ($\geq$4K) scenarios. Recent text-to-image (T2I) diffusion-based Real-ISR approaches \cite{yu2024scaling, yue2024efficient} significantly outperform earlier GAN-based methods \cite{chan2021glean, wang2021towards, ledig2017photo}, owing to the appropriate selection of generative prior and the effective incorporation of these priors into Real-ISR. However, these methods still face limitations; either the selection of pre-trained priors is suboptimal, or the pre-trained image priors are insufficiently exploited. The problem becomes particularly pronounced in ultra-high-resolution scenarios, where patch-wise inference \cite{bar2023multidiffusion} is necessary due to computational constraints. During patch-wise inference, the limited semantic information within each patch conflicts with the global textural guidance derived from the LR image, resulting in undesired artifacts (over-generation). Trivially employing local prompts fails to reconstruct fine textures (under-generation), as such local prompts are inadequate to activate the full generative capacity of the pre-trained model.}

{In this paper, we propose DreamSR, a method tailored for ultra-high-resolution image reconstruction with patch-wise inference, aiming at fully leveraging the pre-trained T2I generative prior. Specifically, we employ FLUX \cite{flux2024} as our backbone due to its unified architecture, which avoids the split of generative priors across separate base and refiner models, i.e., SDXL \cite{podell2023sdxl}.
To address the aforementioned over-generation and under-generation issues, we introduce a dual-branch network architecture Patch Context aware MM-ControlNet. In this design, one ControlNet branch utilizes local prompts to capture visual features of local patches, while the fixed DiT backbone leverages global prompts to fully activate the pre-trained generative prior.}
Based on the proposed architecture, we design a two-stage inference framework consisting of degradation removal and texture generation. We define the multi-step denoising process of the diffusion model as the texture generation stage, which focuses on refining high-frequency textures. 
An accelerated LoRA module is adopted in the degradation removal stage to mitigate the latent distribution shift between degraded LR images and natural images.
The degradation removal stage not only speeds up and reduces the early denoising steps of the reconstruction process, but also aligns the latent space of LR inputs with that of natural images.

To enhance the model's performance on generating fine-grained detail in local patches, we introduce a comprehensive training strategy with stage-specific data processing pipelines to improve the generation ability on local features with limited semantics. For the degradation removal stage, we adopt the traditional Real-ESRGAN \cite{wang2021real} degradation pipeline. For the texture generation stage, we propose a novel image-to-image degradation pipeline to produce training pairs with substantial texture-level discrepancies, thereby improving the model’s capacity for generating fine-grained details with textural guidance. Additionally, a Receptive-Field Enhancement strategy is introduced to align the feature's receptive fields during training and inference, further boosting its overall capability.

The contributions of this work are summarized as follows: (i) We present DreamSR, a multi-stage ISR model architected upon pre-trained DiT models. By leveraging their powerful image priors and our specifically designed MM-DiT network architecture, it effectively pushes the performance ceiling of image super-resolution. (ii) We introduce a Patch Context aware MM-ControlNet, which leverages a dual-branch ControlNet to integrate global textual semantics with local patch-level prompts through cross-attention, enabling alignment between patch-level visual features and local textual information, thereby suppressing over-generation in patch regions.  (iii) We devise a comprehensive training pipeline that utilizes diverse degradation strategies to increase model performance. The incorporation of a Receptive-Field Enhancement mechanism further enhances the model's ability to capture fine-grained details during ultra-high-resolution image reconstruction. (iv) We demonstrate through both quantitative and qualitative evaluations that our DreamSR outperforms state-of-the-art methods on real-world datasets with multiple resolutions.

\section{Related Work}
\begin{figure*}
    \centering
    \includegraphics[width=\linewidth]{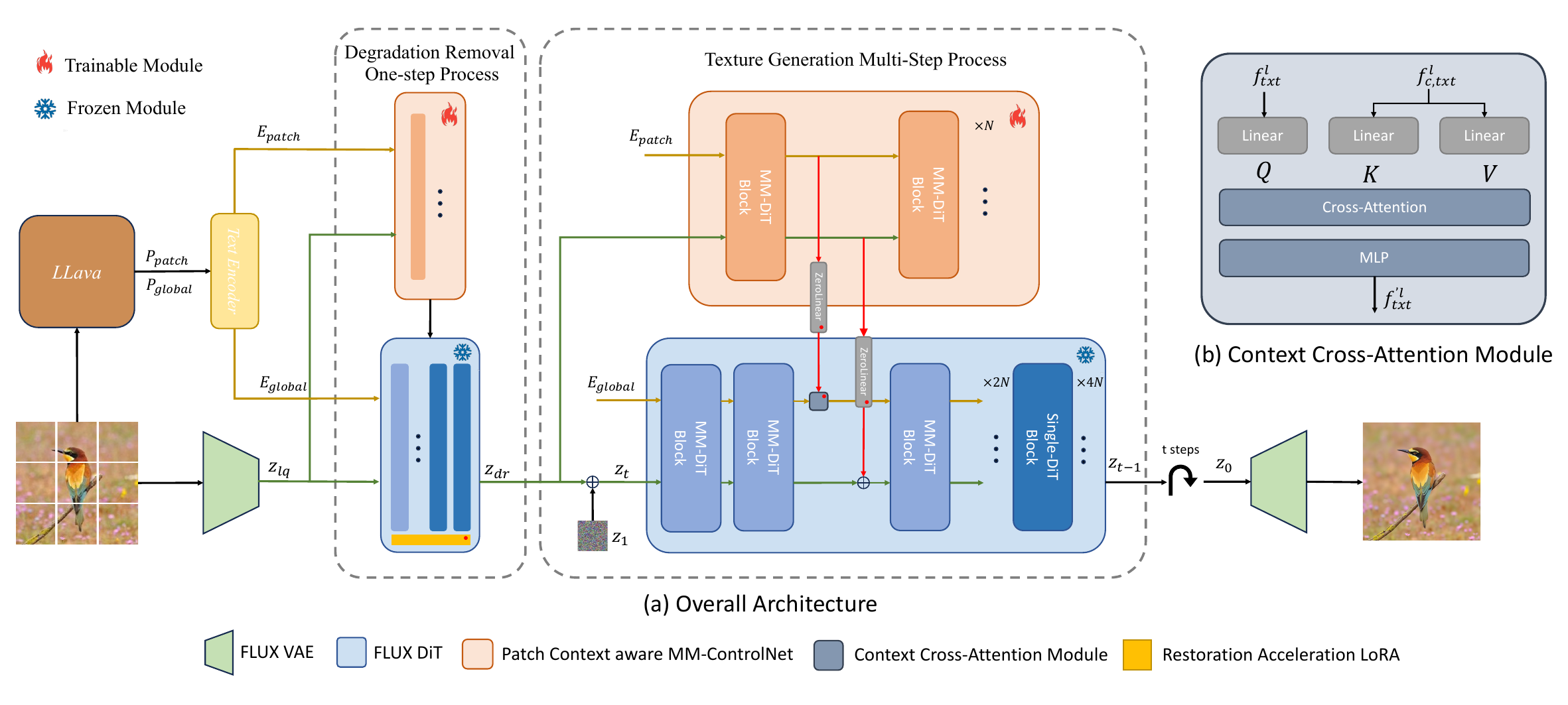}
    \caption{Overview of the proposed DreamSR architecture. Our framework consists of two stages: a degradation removal one-step process and a texture generation process. The Restoration Acceleration LoRA is introduced to compress the early denoising process, while the Patch Context aware MM-ControlNet integrates global and local textual features through the Context Cross-Attention module for effective semantic alignment and fine-grained detail generation.}
    \label{framework}
    \vspace{-2mm}
\end{figure*}

\noindent \textbf{Image Super-Resolution.}
Deep learning-based image super-resolution (ISR) methods have achieved remarkable progress in recent years. CNN-based approaches \cite{dong2014learning, lim2017enhanced, zhang2018image, zhang2018residual, dai2019second} and Transformer-based models \cite{chen2021pre, liang2021swinir, chen2023activating, zhang2022efficient, chen2023dual, liu2023unfolding} have continually advanced the performance upper bound. However, due to the complexity and diversity of real-world degradations, these methods often struggle to effectively remove artifacts and recover fine textures. To mitigate the domain discrepancy between synthetic and real degradations, several studies \cite{wang2021real, zhang2021designing} have focused on learned degradation modeling. Yet, a noticeable gap still remains between synthetically degraded datasets and real-world low-resolution images, leading to suboptimal performance in recovering natural details or enhancing mid-quality real images.

Recently, generative models have been increasingly introduced into ISR to further improve perceptual quality. GAN-based methods \cite{wang2018esrgan,  liang2022details, liang2022efficient, xie2023desra, chen2022real, sun2024coser} significantly enhance the realism of generated details. With the advancement of diffusion models, many approaches \cite{yue2023resshift, cheng2025effective} have integrated them into SR pipelines, achieving impressive results. Several works \cite{wang2024exploiting, yu2024scaling, deng2025acquire, gu2024consissr, qu2024xpsr, wan2024clearsr, xie2024addsr, tsao2024holisdip, liu2025patchscaler} employ ControlNet-like structures to incorporate degradation priors, while others introduce semantic cues \cite{wu2024seesr, yang2024pixel, lin2025harnessing, chen2025adversarial, cui2024taming, li2024distillation, li2025one, noroozi2024you, sun2025pixel, sun2024improving, wang2024sinsr, zhang2024degradation} to exploit the text-image interaction capabilities of text-to-image models, thereby improving detail generation. With the emergence of the Diffusion Transformer (DiT), text–image interaction capabilities have been further strengthened, giving rise to a new wave of DiT-based SR methods. However, these methods generally overlook the crucial alignment between textual information and local image patches at ultra-high resolutions, limiting their ability to fully exploit localized detail generation.

\noindent \textbf{Diffusion Models.}
Diffusion models \cite{ho2020denoising} have demonstrated exceptional capability in generating high-quality and diverse natural images. Latent Diffusion Models (LDMs) \cite{rombach2022high} extend the diffusion process into latent space, greatly improving efficiency while maintaining fidelity in text-to-image synthesis. Through large-scale image–text pretraining, these models effectively capture complex image manifolds and semantic relationships. Several studies \cite{ramesh2022hierarchical, dhariwal2021diffusion, chen2023pixart, saharia2022photorealistic, xie2024sana} have shown that text-to-image diffusion priors outperform GAN-based priors in modeling diverse natural images, motivating their adoption in high-quality super-resolution tasks.
More recently, the Diffusion Transformer (DiT) architecture has emerged as a powerful alternative to conventional U-Net backbones, becoming the new mainstream design. In particular, multi-modal variants such as MM-DiT, used in models like FLUX \cite{flux2024} and SD3.0 \cite{esser2024scaling}, not only scale model capacity but also strengthen image–text interaction, resulting in more semantically aligned visual outputs.  As a result, leveraging such powerful diffusion priors for high-fidelity and controllable image super-resolution has become an active research direction.

\section{Methods}

We build our DreamSR framework upon Flux.1-dev model \cite{flux2024}, aiming to push the performance boundary of image super-resolution by leveraging the powerful pre-trained generative priors of a Diffusion Transformer (DiT). As shown in Figure \ref{framework}, our DreamSR consists of a fixed DiT, a Patch Context aware MM-ControlNet and a Restoration Acceleration LoRA \cite{hu2022lora} module. The reconstruction process is decomposed into two stages: degradation removal and texture generation. In the degradation removal stage, the Restoration Acceleration LoRA module is integrated into the DiT network, which compresses the early-stage denoising process and reduces inference computational cost. In the texture generation stage, the diffusion model (DiT $\&$ ControlNet) performs multi-step denoising to synthesize and refine high-frequency textures.

To mitigate the over-generation artifacts that arise from semantic mismatches between local patches and global text during patch-wise inference, we propose a Patch Context aware MM-ControlNet. Unlike previous methods \cite{ai2024dreamclear} that inject only visual features from ControlNet into DiT, our approach integrates both visual and textual features from the MM-DiT blocks , where local textual information is seamlessly incorporated and effectively combined with global semantics. This design facilitates consistent semantic understanding across patches, ensuring coherent global structures and precise local texture reconstruction.

\subsection{Patch Context aware MM-ControlNet}

We construct our Patch Context aware MM-ControlNet $\mathcal{F}_\theta$ by partially copying the layers from the DiT network $\epsilon_\phi$ in FLUX.1-dev \cite{flux2024}. 
To enhance the text interaction capability of $\mathcal{F}_\theta$, we selectively employ these text-related blocks (MM-DiT) while discarding Single DiT blocks. These text-related blocks can efficiently process patch-level textual information and integrate the resulting features into the main network, strengthening its overall capability. Furthermore, to balance restoration performance and inference efficiency, we perform a trimmed replication of the MM-DiT by keeping only half of the MM-DiT blocks in $\mathcal{F}_\theta$. The intermediate features from these blocks are {evenly injected} into the corresponding MM-DiT blocks of the main network.

During inference, given an LQ image $I_{lq}\in \mathbb{R}^{H \times W \times C}$,  we first partition it into overlapping patches $I_{\text{patch}} = \left\{ I_{ij} \in \mathbb{R}^{p \times p \times C} \right\}$ for subsequent patch-wise processing. We use a multi-modal LLM LLaVA \cite{liu2023visual} to generate both a global prompt $P_{global}$ describing the entire image and localized prompts $P_{patch}=\{P_{ij}\}$  corresponding to individual image patches. During network processing, $I_{lq}$ is first processed by Encoder $\mathcal{E}$, then gets the latent present $\textbf{z}_{lq}$. 
The patch latent $\textbf{z}^{i,j}_{lq}$ partitioned from $\textbf{z}_{lq}$ and local prompt embedding $E_{ij}$ are fed into the MM-ControlNet, while noisy patch latent $\textbf{z}^{i,j}_t$ partitioned from $\textbf{z}_t$ and global text prompt embedding $E_{global}$ are injected directly into the DiT network. For a patch latent $\textbf{z}^{ij}_{lq}$, the MM-ControlNet outputs a series image features $f_{c,img} = \{ f^{l}_{c,img}\}$ and text features $f_{c,txt}=  \{ f^{l}_{c,txt}\},l=1,...,9$. For each block $l$, as for the image features, a zero-MLP is used and the feature is directly injected into the main DiT block as
\begin{equation}
  f'^{l}_{img} = f^l_{img} + \mathcal{M}^l(f^l_{c, img}),
  \label{eq:important}
\end{equation}
where $\mathcal{M}^l_{img}(.)$ is the zero-MLP layer. As for text features, we introduce a Context Cross-Attention module to enable effective fusion of global and local textual information, which can be defined as
\begin{equation}
  f'^l_{txt} = \mathcal{M}_{txt}^{l}(CrossAttention(Q^l_{txt},K^l_{c,txt},V^l_{c,txt})),
  \label{eq:important}
\end{equation}
where $Q^{l}_{txt}=P_{Q}(f^{l}_{txt})$,$K^{l}_{c,txt}=P_{K}(f^{l}_{c,txt})$ and $V^{l}_{c,txt}=P_{V}(f^{l}_{c,txt})$,$\mathcal{M}^l_{txt}(.)$ is the MLP layer and $P(\cdot)$  represents a linear projection function.
This cross-attention mechanism enables dynamic emphasis on the most relevant semantic cues from the global prompt for each local patch while suppressing irrelevant or redundant contextual signals. Consequently, our design facilitates more precise details generation within each local region and preserves the semantic consistency of the overall image. 

After obtaining the predicted velocity field $v^{ij}_p$ for each patch latent $\textbf{z}^{ij}_t$ from the network, we follow the same procedure as Multi-Diffusion \cite{bar2023multidiffusion} to aggregate all patch outputs and get $v_p$ , thereby enhancing the global semantic coherence and consistency of the reconstructed image.

\subsection{Restoration Acceleration LoRA}
Based on the designed model architecture, to increase efficiency when reconstructing high-resolution images, we define a two-stage inference process: degradation removal and texture generation. Many recent diffusion-based SR methods \cite{lin2024diffbir, ai2024dreamclear, yu2024scaling, yang2024pixel} adopt a degradation removal module { to prevent the network from confusing image semantics with degradation information}{, allowing the main network to focus more on details generation rather than removing image-independent degradation patterns}, thereby improving the generative quality and stability of the reconstruction. Inspired by one-step diffusion SR approaches \cite{wu2024one,dong2025tsd}, we elegantly integrate the degradation removal process into our network, enabling rapid degradation elimination through a single forward inference while effectively compressing the overall inference process. The restored latent produced in this stage provides a cleaner and more semantically consistent representation, which significantly benefits the subsequent multi-step texture generation phase.

Specifically, during the first inference step, following common practices of the one-step SR model, we augment the DiT network with a Restoration Accelerate LoRA module. The LQ latent $\textbf{z}_{lq}$ is fed into both the main DiT network and the MM-ControlNet for a single forward pass, producing latent output with degradations removed $\textbf{z}_{dr}$. $\textbf{z}_{dr}$ serves as the condition input for the MM-ControlNet.
After getting degradation-removed results, unlike previous methods that still initiate the multi-step generation from pure noise, we utilize $\textbf{z}_{dr}$ as the intermediate state of the diffusion process. Specifically, the noisy latent $\mathbf{z}_{t}$ is abtained by injecting controlled Gaussian noise into $\textbf{z}_{dr}$ as

\begin{equation}
    \mathbf{z}_{t} = (1 - t)\mathbf{z}_{dr} + t \cdot \boldsymbol{\epsilon},
\end{equation}
where $t$ is the predefined interpolation step. Starting the diffusion process from this intermediate state $\mathbf{z}_{t}$ allows the model to bypass redundant early-stage denoising iterations and instead focus on fine-grained texture refinement, efficiently generating high-quality results.

\subsection{Training strategy}
\label{sec:3.3}
\begin{figure}
    \centering
    \includegraphics[width=\linewidth]{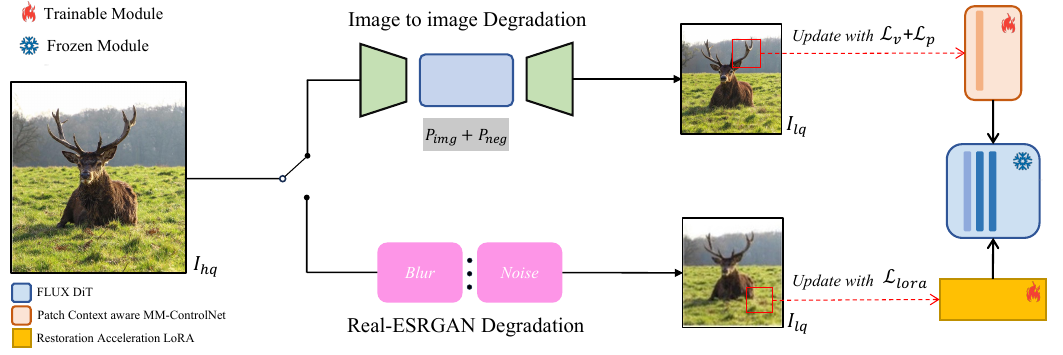}
    \caption{Overall training pipeline for DreamSR.}
    \vspace{-4mm}
    \label{i2i}
\end{figure}

To address the insufficient detail generation issue, we develop a comprehensive training strategy with stage-specific data processing pipelines, enhancing the model’s sensitivity to local textural guidance. We also develop Receptive-Field Enhancement strategy to align the model’s receptive fields during training and inference, boosting its overall capability. The overall training pipeline is shown in Figure \ref{i2i}.

\noindent \textbf{Stage-specific degradation pipelines.} {Since the two model stages have distinct priorities, we employ different degradation pipelines to generate training pairs. For the first stage (degradation removal), we use the traditional Real-ESRGAN pipeline \cite{wang2021real}. For the second stage (texture generation), we introduce an image-to-image (i2i) degradation method that leverages the FLUX model to erase details from high-quality images, enhancing detail generation through textural guidance. Unlike Real-ESRGAN’s pixel-wise degradation, which compromises both structure and texture, our i2i approach selectively removes texture details while maintaining global structural consistency. This allows the network to focus on reconstructing high-frequency details with textual guidance, improving output fidelity and realism.  Specifically, we start with a high-quality image $I_{hq}$, an image prompt $P_{img}$ and a negative prompt $P_{neg}$. After downsampling $I_{hq}$, we encode it via a VAE encoder to obtain $\textbf{z}_{hq}$. This latent representation is blended with Gaussian noise (strength: 0.3–0.5) and then partially denoised using the FLUX model conditioned on $P_{img} + P_{neg}$.}

\noindent \textbf{Joint training.} During training, we employ a joint optimization strategy to concurrently train the proposed MM-ControlNet and the Restoration Accelerate LoRA module. Specifically, in the initial phase, we first optimize the MM-ControlNet using the datasets generated by the i2i degradation pipeline. The initial training process can stabilize the output. The model typically predicts velocity field $v_p$ and we use velocity loss to update the MM-ControlNet parameters. The velocity loss can be formulated as 
\begin{equation}
  {v}_{gt} = \textbf{z}_1 - \textbf{z}_0,  \label{eq:important}
\end{equation}
\begin{equation}
  {\mathcal{L}}_{v} = ||v_p - v_{gt}||^2_2,  \label{eq:important}
\end{equation}
where $\textbf{z}_0=\mathcal{E}(I_{gt})$ is the target latent, $\textbf{z}_1$ is the added random noise. In addition, we introduce pixel-level loss to further enhance the reconstruction fidelity. When the sampled $t$ in range $0\sim0.2$, we get the prediction output $\hat{\textbf{z}}_0=\textbf{z}_t-tv_p$ and apply it with the VAE Decoder $\mathcal{D}$ into pixel space. Besides ${\mathcal{L}}_{v}$, we additionally use MSE loss and perceptual loss  \cite{zhang2018unreasonable} to update the parameter. The pixel loss can be defined as
\begin{equation}
  {\mathcal{L}}_{p} = ||\mathcal{D}(\hat{\textbf{z}}_0) - I_{gt}||^2_2 + \lambda LPIPS(\mathcal{D}(\hat{\textbf{z}}_0),I_{gt}),  \label{eq:important}
\end{equation}
where $\lambda$ is a weighting scalar. 

Subsequently, during the iterative training process, the network to be updated is randomly selected at each step. 
When training the Restoration Acceleration LoRA module for the one-step degradation removal stage, we modify the training objective to image-level loss, directly supervising the model with the ground-truth image. To further enhance perceptual quality, we employ a combination of perceptual loss and GAN loss. Additionally, since the latent output from the one-step process is directly fed into the subsequent multi-step generation stage, a latent-level loss is introduced to refine the learned representations and ensure consistency across the two stages. The whole loss can be written as
\begin{equation}
\begin{split}
\mathcal{L}_{lora} =& ||\hat{\textbf{z}}_{dr} - {\textbf{z}}_0 ||^2_2  + ||\mathcal{D}(\hat{\textbf{z}}_{dr}) - I_{gt}||^2_2 \\
& + \lambda_1LPIPS(\mathcal{D}(\hat{\textbf{z}}_{dr}),I_{gt}) +\lambda_2GAN(\mathcal{D}(\hat{\textbf{z}}_{dr})),
\end{split}
\end{equation}
where $\lambda_1$ and $\lambda_2$ are weighting scalars. The detailed training schemes are provided in the supplementary material.

\noindent \textbf{Receptive-Field Enhancement training.} {To enable ultra-high-resolution image restoration, it is crucial to align the data distributions between training and inference. The key lies in the distinction between local and global prompts. Unlike previous approaches that rely on downsampling or randomly cropping large patches (e.g., 768×768 or 1024×1024) from medium-resolution (around 1K resolution) datasets, e.g., LSDIR \cite{li2023lsdir} and FFHQ \cite{karras2019style}, where local prompts become too similar to global prompts, causing the network to overlook local prompts guidance during reconstruction. So we introduce a Receptive-Field Enhancement training strategy, ensuring sufficient divergence between local and global prompts, thereby reinforcing the importance of local prompt conditions in the reconstruction process. Specifically, we directly extract 512×512 patches from native high-resolution images (mostly around 2K resolution).}
\section{Experiments}

\begin{table*}[t]
\footnotesize
\setlength{\tabcolsep}{1.5pt}
\centering
\caption{Quantitative comparison with state-of-the-art methods on real-world image super-resolution benchmarks. The best and second performance are highlighted in {\color{red}red} and {\color{blue}blue}, respectively.}
\vspace{-1mm}
\resizebox{\textwidth}{!}{
\begin{tabular}{c|c|cc|ccccccccc}
\toprule
\multirow{2}{*}{Datasets} & \multirow{2}{*}{Metrics} & \multicolumn{10}{c}{Methods} \\
\cmidrule{3-13}
& & {\scriptsize BSRGAN\cite{zhang2021designing}} & {\scriptsize Real-ESRGAN\cite{wang2021real}} & {\scriptsize StableSR\cite{wang2024exploiting}} & {\scriptsize DiffBIR\cite{lin2024diffbir}} & {\scriptsize SeeSR\cite{wu2024seesr}} & {\scriptsize SUPIR\cite{yu2024scaling}} & {\scriptsize OSEDiff\cite{wu2024one}} & {\scriptsize FaithDiff\cite{chen2025faithdiff}} & {\scriptsize DreamClear\cite{ai2024dreamclear}} & {\scriptsize DiT4SR\cite{duan2025dit4sr}} & {\scriptsize Ours} \\
\midrule
\multirow{6}{*}{RealSR}
& PSNR $\uparrow$ & \color{red}{26.3785} & \color{blue}{25.6855} & 25.5091 & 24.8340 & 25.1465 & 22.4082 & 23.3748 & 24.5681 & 25.0070 & 23.5397 & 20.9305 \\
& SSIM $\uparrow$ & \color{red}0.7651 & \color{blue}0.7614 & 0.7491 & 0.6501 & 0.7210 & 0.5995 & 0.7105 & 0.6927 & 0.7026 & 0.6680 & 0.6058 \\
& LPIPS $\downarrow$ & \color{blue}0.2656 & 0.2710 & \color{red}0.2604 & 0.3650 & 0.3008 & 0.4204 & 0.3032 & 0.2949 & 0.3258 & 0.3143 & 0.4079 \\
& MUSIQ $\uparrow$ & 63.2871 & 60.3671 & 61.8059 & 69.2790 & \color{blue}69.8241 & 63.4065 & 69.2206 & 68.7046 & 58.3191 & 67.5601 & \color{red}70.5560 \\
& MANIQA $\uparrow$ & 0.3758 & 0.3737 & 0.3769 & \color{blue}0.5582 & 0.5437 & 0.4472 & 0.4847 & 0.4742 & 0.4313 & 0.4540 & \color{red}{0.5731} \\
& CLIPIQA+ $\uparrow$ & 0.5822 & 0.5840 & 0.6145 & \color{blue}0.7233 & 0.6912 & 0.5989 & 0.7200 & 0.6720 & 0.5920 & 0.6609 & \color{red}{0.7318} \\
\midrule
\multirow{6}{*}{DRealSR}
& PSNR $\uparrow$ & \color{blue}28.7025 & 28.6147 & \color{red}{29.0182} & 25.9035 & 28.0707 & 24.4641 & 25.9906 & 26.2201 & 28.5904 & 25.6479 & 24.0158 \\
& SSIM $\uparrow$ & 0.8028 & \color{red}0.8052 & \color{blue}{0.8044} & 0.6245 & 0.7683 & 0.6198 & 0.7475 & 0.6857 & 0.7532 & 0.6772 & 0.6484 \\
& LPIPS $\downarrow$ & \color{blue}0.2858 & \color{red}0.2819 & 0.2898 & 0.4669 & 0.3175 & 0.4650 & 0.3138 & 0.3669 & 0.3680 & 0.3745 & 0.4639 \\
& MUSIQ $\uparrow$ & 57.1687 & 54.2752 & 51.3571 & 66.1187 & 65.0895 & 62.3864 & 65.2829 & \color{blue}{66.5850} & 43.4771 & 65.1260 & \color{red}{67.3315} \\
& MANIQA $\uparrow$ & 0.3403 & 0.3440 & 0.3177 & \color{red}0.5543 & 0.5129 & 0.4542 & 0.4765 & 0.4575 & 0.3055 & 0.4425 & \color{blue}0.5404 \\
& CLIPIQA+ $\uparrow$ & 0.5621 & 0.5546 & 0.5349 & \color{red}0.7173 & 0.6792 & 0.6145 & 0.7008 & 0.6686 & 0.5060 & 0.6668 & \color{blue}0.7028 \\
\midrule
\multirow{3}{*}{RealLQ250}
& MUSIQ $\uparrow$ & 63.5230 & 62.5185 & 56.6588 & 67.5315 & 70.3712 & 65.9080 & 69.2500 & 69.6166 & 67.1254 & \color{red}{71.8311} & \color{blue}{71.6516} \\
& MANIQA $\uparrow$ & 0.3515 & 0.3565 & 0.2932 & 0.4900 & 0.4894 & 0.3848 & 0.4245 & 0.3971 & \color{blue}{0.4938} & 0.4607 & \color{red}{0.5319} \\
& CLIPIQA+ $\uparrow$ & 0.6019 & 0.6116 & 0.5709 & 0.6917 & 0.7037 & 0.6542 & 0.7077 & 0.6792 & 0.6922 & \color{blue}{0.7106} & \color{red}{0.7206} \\
\midrule
\multirow{3}{*}{RealLR200}
& MUSIQ $\uparrow$ & 64.8724 & 62.9628 & 63.2798 & 66.4224 & 69.6413 & 63.8806 & 69.2590 & 68.7590 & 66.3161 & \color{blue}70.6282 & \color{red}{70.6862} \\
& MANIQA $\uparrow$ & 0.3704 & 0.3689 & 0.3689 & 0.4736 & 0.4698 & 0.3979 & 0.4426 & 0.4268 & \color{blue}{0.4758} & 0.4650 & \color{red}{0.5488} \\
& CLIPIQA+ $\uparrow$ & 0.6260 & 0.6221 & 0.6409 & 0.6918 & \color{blue}{0.7088} & 0.6418 & 0.7149 & 0.6846 & 0.6912 & 0.7086 & \color{red}{0.7391} \\
\midrule
\multirow{3}{*}{RealDeg}
& MUSIQ $\uparrow$ & 43.8358 & 44.8099 & 35.5875 & 40.6098 & \color{blue}45.5557 & 45.5267 & 44.6909 & 47.5436 & 41.8048 & 44.7736 & \color{red}45.6234  \\
& MANIQA $\uparrow$ & 0.3648 & 0.3848 & 0.3291 & 0.4413 & \color{blue}0.4535 & 0.3613 & 0.4391 & 0.3862 & 0.3854 & 0.4437 & \color{red}{0.4574}  \\
& CLIPIQA+ $\uparrow$ & 0.5921 & 0.6102 & 0.5358 & 0.6367 & 0.6579 & 0.6094 & 0.6534 & 0.6329 & 0.6290 & \color{red}{0.6841} & \color{blue}0.6647  \\
\bottomrule

\end{tabular}
}
\vspace{-3mm}
\label{tab:image_quality}
\end{table*}

\subsection{Datasets and implementation details}
\textbf{Datasets.} For training, we utilize a collected dataset consisting of 580,000 high-quality general images and 120,000 face images, each paired with corresponding text descriptions. We pre-generate the corresponding LR counterparts using our i2i degradation pipeline, while the Real-ESRGAN degradation method is also applied online during the training process with the same configuration as SeeSR \cite{wu2024seesr}.

Since our method specifically focuses on the Real-ISR task, we evaluate our model on widely used real-world datasets, including RealSR \cite{cai2019toward}, DrealSR \cite{wei2020component}, RealLR200 \cite{wu2024seesr}, RealLQ250 \cite{ai2024dreamclear} and RealDeg \cite{chen2025faithdiff} social media subset. Following SeeSR, the LR images from RealSR and DrealSR are center-cropped to $128\times128$. RealLR200 contains images ranging from $100\times100$ to $400\times400$, RealLQ250 provides images at $256\times256$, and RealDeg includes high-resolution images with a short edge of 720. All experiments are conducted with the scaling factor of $\times4$.

\noindent \textbf{Evaluation Metrics.}
In order to comprehensively evaluate the performance of different methods, we employ a suite of both reference-based and no-reference metrics. For reference-based assessment, we utilize PSNR and SSIM \cite{wang2004image}  to measure reconstruction fidelity, and LPIPS \cite{zhang2018unreasonable} to quantify perceptual quality. For no-reference evaluation, we include MANIQA \cite{yang2022maniqa}, MUSIQ \cite{ke2021musiq}, and CLIPIQA+ \cite{wang2023exploring} to assess image quality from diverse perspectives.

\noindent \textbf{Implementation details.} Our model is trained on 16 H20 GPUs with a batch size of 64 using AdamW optimizer (lr=5e-6). During training, images are randomly cropped to $512\times512$ patches with online-generated local text prompts. The rank of Restoration Acceleration LoRA is set to 256. 
{We first train the MM-ControlNet alone for 20,000 steps to stabilize the output. Subsequently, the model alternates between a multi-step texture generation stage and a single-step degradation removal stage for approximately 100,000 steps. The hyperparameters $\lambda$, $\lambda_1$, and $\lambda_2$ are set to 2, 2 and 0.5, respectively.} To support both patch-based high-resolution restoration and full-image SR for smaller inputs, we resize images to a 1024-pixel shorter side with 0.2 probability while maintaining local text consistency with global text. During inference, we set $t=0.8$ and the total number of steps to 17, where the first step performs degradation removal and the remaining 16 steps are used for texture generation, resulting in an overall 1+16 inference scheme.  

\begin{figure*}
    \centering
    \vspace{-2mm}
    \includegraphics[width=\linewidth]{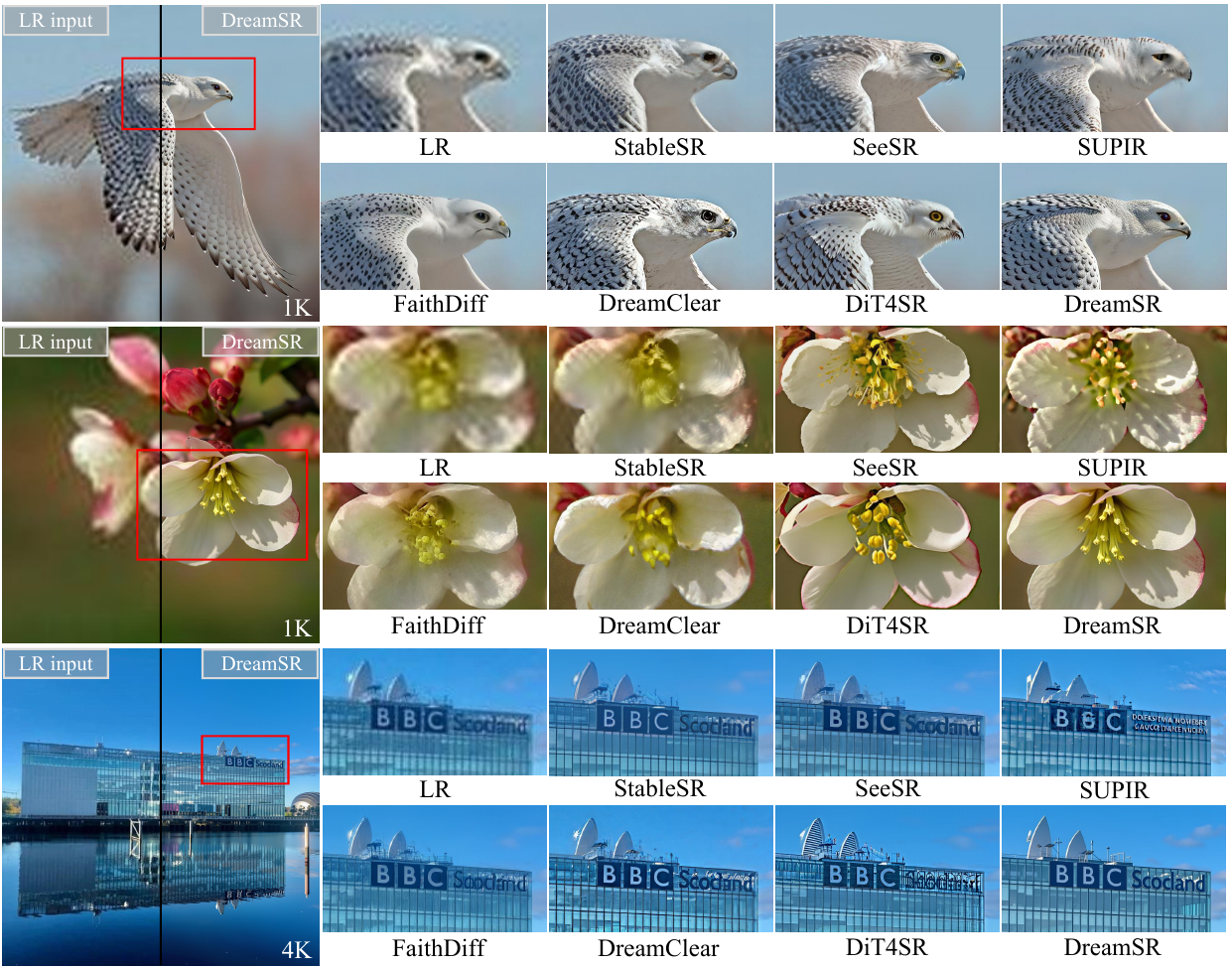}
    
    \caption{Qualitative comparisons with different methods on real-world datasets. Our DreamSR achieves the best performance, generating more realistic images with fine-scale structures and details. More visual results are included in the supplementary material.}
    \label{results}
    \vspace{-2mm}
\end{figure*}

\subsection{Comparison with Existing Methods}
\label{sec:4.2}

We compare our method with state-of-the-art Real-ISR methods, including GAN-based methods (i.e. BSRGAN \cite{zhang2021designing}, Real-ESRGAN \cite{wang2021real}) and diffusion-based methods(i.e. StableSR \cite{wang2024exploiting}, DiffBIR \cite{lin2024diffbir}, SeeSR \cite{wu2024seesr}, SUPIR \cite{yu2024scaling}, OSEDiff \cite{wu2024one}, FaithDiff \cite{chen2025faithdiff}, DreamClear \cite{ai2024dreamclear}, DiT4SR \cite{duan2025dit4sr}). We use the publicly released codes and models, with default inference configurations for testing. 

\noindent \textbf{Quantitative Comparisons.} We first show the quantitative comparison on the five real-world datasets in Table \ref{tab:image_quality}. It can be seen that our method achieves competitive performance on 5 datasets containing various resolutions. For the $128\times128$ resolution RealSR and DRealSR datasets, our method reaches the best results on non-reference metrics, which reflects the excellent image quality of our results. Due to the inherently strong detail generation capability of diffusion-based models, none of the diffusion-based approaches, including ours, achieves outstanding performance on reference metrics. This phenomenon has also been widely noted in prior studies \cite{yu2024scaling, jinjin2020pipal, blau2018perception} that existing reference metrics are not suitable for evaluation with the model's generative ability increasing. On the RealLQ250 and RealLR200 datasets, our method exhibits overwhelming performance, consistently achieving top results across all non-reference metrics. Furthermore, on the high-resolution RealDeg dataset, our model demonstrates excellent objective performance, validating the superiority of our approach in handling high-resolution image reconstruction tasks.

\noindent \textbf{Qualitative Comparisons.} The qualitative comparisons
against other methods are provided in Figure \ref{results}, which clearly demonstrate that our approach generates more natural textures and preserves fine-grained details better than existing methods. For the first two rows, which are sampled from the RealLQ250 dataset with $256\times256$ low-resolution inputs, our method reconstructs clearer and more plausible details, achieving visually superior results. For the last row, which is selected from the RealDeg dataset with high-resolution outputs, we observe that many competing methods fail to properly recover consistent details due to semantic misalignment during inference, often generating distorted or incorrect textures, while some approaches are unable to produce fine-grained details, leading to overly smooth reconstructions. In contrast, our approach produces structurally coherent and detailed reconstructions, demonstrating its robustness and superior visual fidelity.

\begin{table}[!t]
\small
\caption{Inference-time performance of diffusion-based SR methods for the diffusion process on 2560x1440 resolution images. All methods are tested on a H20 GPU. The best and second performances are highlighted in {\color{red}red} and {\color{blue}blue}, respectively.}
\centering
\vspace{-2mm}
\setlength{\tabcolsep}{2pt}
\resizebox{1.0\columnwidth}{!}{
\begin{tabular}{ccccccccc}
\toprule
                  & StableSR & DiffBIR & SeeSR & SUPIR & FaithDiff & DreamClear & DiT4SR  &Ours \\ \bottomrule
Step    & 200      & 50   & 50    &  50   &  \color{blue}20  & 50 & 40 & \color{red}{1+16}  \\
Time (s) & 156  & 127 & 119 &  124  &  \color{red}30 & 185 & 228 &  \color{blue}86 \\ \bottomrule
\end{tabular}}
\vspace{-6mm}
\label{tab:complexity}
\end{table}

\noindent \textbf{Inference time Comparisons.}
We evaluate the inference efficiency of our model on standard 2K images using a single H20 GPU. Benefiting from our acceleration design, the proposed method achieves high-quality detail reconstruction with only 1+16 inference steps. Our step count is significantly lower than that of other diffusion-based SR methods. Despite the relatively large parameter size of the DiT architecture, our model still runs faster than most methods built upon SD \cite{rombach2022high} or SDXL \cite{podell2023sdxl} UNet backbones, and is only slower than FaithDiff, which adopts a lightweight design. Overall, our approach achieves a favorable balance between inference speed and reconstruction quality.

\subsection{Ablation Study}

\begin{figure}
    \centering
    \vspace{-3mm}
    \includegraphics[width=\linewidth]{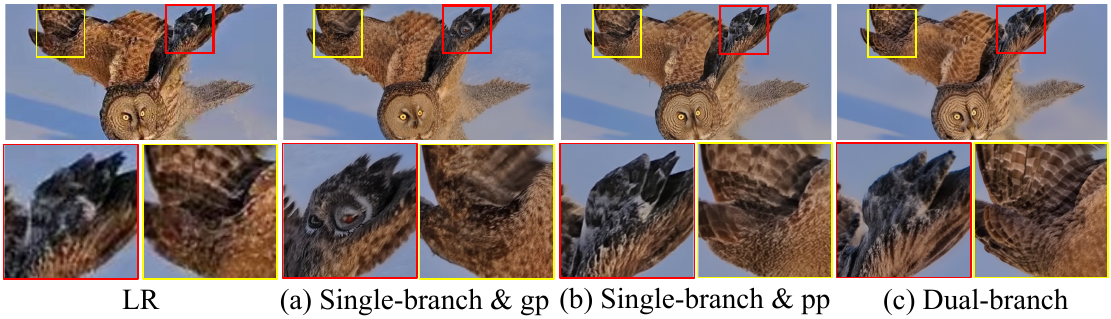}
    \vspace{-6mm}
    \caption{Visual comparison of different prompt configurations.
(a) Single-branch ControlNet with global prompt, (b) single-branch ControlNet with patch-level prompt, and (c) our full dual-branch model.}
    \label{results_ab_prompt}
    \vspace{-5mm}
\end{figure}

\noindent \textbf{Effectiveness of dual-branch MM-ControlNet.}
To validate the effectiveness of the proposed dual-branch MM-ControlNet, we conduct an ablation study by replacing our MM-ControlNet with a conventional ControlNet architecture that removes the text feature branch. The model is trained under the same settings as the full model, except that it uses global prompt. During testing, we evaluate two variants: one using only the global prompt and the other using only patch-level prompts as botch DiT and ControlNet inputs. We evaluate the performance on the RealDeg dataset.
As shown in Table \ref{ab_prompt}, both variants perform worse than the full version of DreamSR across all metrics. Moreover, qualitative results shown in Figure \ref{results_ab_prompt} reveal that when conditioned solely on the global prompt, the model tends to suffer from over-generation artifacts. Conversely, when conditioned only on patch-level prompts, the model lacks sufficient global semantic awareness, leading to unnatural and incoherent textures. These observations demonstrate that our dual-branch ControlNet design effectively enables interaction between global and local textual information through cross-attention, thereby enhancing both semantic consistency and detail generation in reconstructed images.

\begin{table}[t]
\footnotesize
\renewcommand{\arraystretch}{1.0}
\setlength{\tabcolsep}{2.5pt}
\centering
\caption{Ablation results of different architectures of ControlNet on RealDeg dataset. After removing the text feature branch of ControlNet, we use different prompt settings to test.}
\vspace{-2mm}
\resizebox{1.0\columnwidth}{!}{
\begin{tabular}{ccc|cccc}
\toprule
\scriptsize{Text branch} & \scriptsize{{Global prompt}} & \scriptsize{Patch prompt} & \scriptsize{MUSIQ $\uparrow$} & \scriptsize{MANIQA $\uparrow$} & \scriptsize{CLIPIQA+$\uparrow$}\\
\midrule

w/o&$\checkmark$ & & 41.6816 & 0.3891 & 0.6251 \\
w/o& & $\checkmark$ & 41.5523 & 0.3932 & 0.6177 \\
w&$\checkmark$ & $\checkmark$ & 45.6234 & 0.4574 & 0.6647 \\
\bottomrule
\end{tabular}
}
\label{ab_prompt}
\end{table}

\begin{figure}
    \centering
    \vspace{-2mm}
    \includegraphics[width=\linewidth]{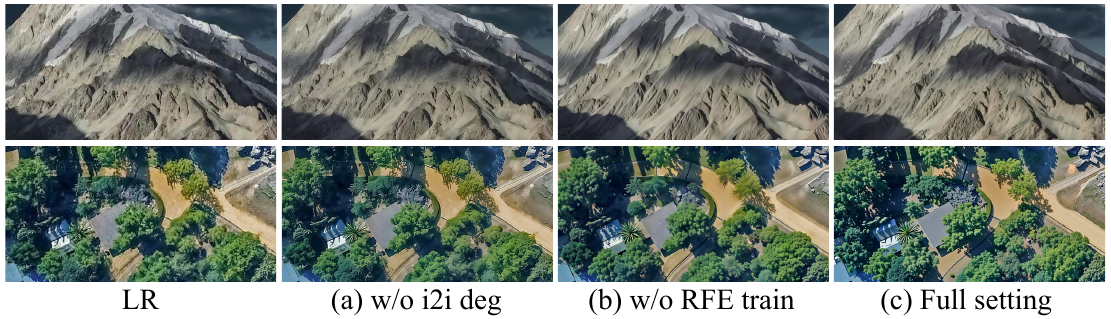}
    \vspace{-7mm}
    \caption{Visual comparison of diffeterent training strategies, (a) Without the proposed i2i degradation, (b) without receptive-field enhancement training, and (c) our full model.}
    \label{results_ab_train}
    \vspace{-3mm}
\end{figure}

\noindent \textbf{Effectiveness of Training Strategy.} To verify the effectiveness of our proposed training strategy, which includes diverse degradation modeling and Receptive-Field Enhancement training, we conduct ablation experiments on the RealDeg dataset. Specifically, we test two simplified settings: (1) training with only Real-ESRGAN degradation, and (2) removing receptive-field enhancement strategy by resizing the short side of images to 1024 and cropping 1024×1024 patches for training. As shown in Table \ref{ab_train}, both ablated settings result in a clear drop across all non-reference metrics, confirming the effectiveness of our full training design. 
As shown in Figure \ref{results_ab_train}, when only Real-ESRGAN degradation is used, the model fails to reconstruct rich image details resulting in overly smooth and texture-lacking outputs. In contrast, our i2i degradation preserves global structures while introducing stronger local detail variations, effectively improving texture synthesis. Moreover, without the receptive-field enhancement strategy, the model tends to overfit global semantics and leads to degraded reconstruction of local details. These results demonstrate that both training components are crucial for achieving high-quality detail generation in ultra-high-resolution restoration.

\begin{table}[t]
\scriptsize
\renewcommand{\arraystretch}{1.0}
\setlength{\tabcolsep}{4.0pt}
\centering
\caption{Ablation results of diffeterent training strategies on RealDeg dataset.}
\vspace{-2mm}
\begin{tabular}{c|cccc}
\toprule
 {Setting} & {MUSIQ $\uparrow$} & {MANIQA $\uparrow$} & {CLIPIQA+$\uparrow$}\\
\midrule

w/o i2i deg & 43.7706 & 0.3917 & 0.6355\\
w/o RFE train & 43.5546 & 0.3868 & 0.5975 \\
Full setting & 45.6234 & 0.4574 & 0.6647\\
\bottomrule
\end{tabular}
\label{ab_train}

\end{table}

\begin{table}[t]
\scriptsize
\renewcommand{\arraystretch}{1.0}
\setlength{\tabcolsep}{3.5pt}
\centering
\caption{Ablation results of Restoration Acceleration LoRA on RealDeg dataset. All models are trained using the Real-ESRGAN degradation for fair comparison.}
\vspace{-3mm}
\begin{tabular}{cc|cccc}
\toprule
{RA LoRA} & {Inference steps} & {MUSIQ $\uparrow$} & {MANIQA $\uparrow$} & {CLIPIQA+$\uparrow$}\\
\midrule

w/o & 20  & 37.9976 & 0.3868 & 0.5975 \\
w/o & 30 & 38.3402 & 0.3865 & 0.6042 \\
w & 1+16 & 43.7706 & 0.3917 & 0.6355 \\
\bottomrule
\end{tabular}
\vspace{-5mm}
\label{ab_lora}
\end{table}

\noindent \textbf{Effectiveness of Restoration Acceleration LoRA.} To evaluate the effectiveness of the proposed Restoration Acceleration LoRA, we conduct an ablation study by removing the LoRA module and training the model only on the multi-step diffusion stage. It is worth noting that, since different degradation strategies may introduce additional variables affecting the LoRA’s performance, we adopt only the Real-ESRGAN degradation setting for this experiment to ensure a fair comparison. As shown in Table \ref{ab_lora}, when the LoRA module is removed, even with a larger number of inference steps, the performance consistently lags behind our full multi-stage model across all metrics. This result demonstrates that the proposed LoRA effectively compresses the early-stage denoising process without compromising restoration quality, validating the rationality and efficiency of our two-stage inference framework.

\section{Conclusion}

In this paper, we introduced DreamSR, a diffusion-based framework for real-world ultra-high-resolution image super-resolution. By addressing the semantic misalignment inherent in patch-wise inference, DreamSR effectively suppresses local over-generation and enhances fine-detail synthesis. Our dual-branch MM-ControlNet integrates global and local textual guidance through cross-attention, achieving consistent semantic alignment between patches and global context. Furthermore, the proposed Receptive Field Enhancement training strategy enables the model to better capture local texture information while maintaining global structural fidelity. Extensive experiments demonstrate that DreamSR achieves state-of-the-art performance and superior perceptual quality on real-world benchmarks.
{
    \small
    \bibliographystyle{ieeenat_fullname}
    \bibliography{main}

@String(ECCV= {Eur. Conf. Comput. Vis.})

@String(ICLR = {Int. Conf. Learn. Represent.})

@String(AAAI = {AAAI})

@String(ECCV  = {ECCV})

@String(ICLR  = {ICLR})

@inproceedings{dong2014learning,
  title={Learning a deep convolutional network for image super-resolution},
  author={Dong, Chao and Loy, Chen Change and He, Kaiming and Tang, Xiaoou},
  booktitle={European conference on computer vision},
  pages={184--199},
  year={2014},
  organization={Springer}
}

@inproceedings{lim2017enhanced,
  title={Enhanced deep residual networks for single image super-resolution},
  author={Lim, Bee and Son, Sanghyun and Kim, Heewon and Nah, Seungjun and Mu Lee, Kyoung},
  booktitle={Proceedings of the IEEE conference on computer vision and pattern recognition workshops},
  pages={136--144},
  year={2017}
}

@inproceedings{zhang2018residual,
  title={Residual dense network for image super-resolution},
  author={Zhang, Yulun and Tian, Yapeng and Kong, Yu and Zhong, Bineng and Fu, Yun},
  booktitle={Proceedings of the IEEE conference on computer vision and pattern recognition},
  pages={2472--2481},
  year={2018}
}

@inproceedings{zhang2018image,
  title={Image super-resolution using very deep residual channel attention networks},
  author={Zhang, Yulun and Li, Kunpeng and Li, Kai and Wang, Lichen and Zhong, Bineng and Fu, Yun},
  booktitle={Proceedings of the European conference on computer vision (ECCV)},
  pages={286--301},
  year={2018}
}

@inproceedings{liang2021swinir,
  title={Swinir: Image restoration using swin transformer},
  author={Liang, Jingyun and Cao, Jiezhang and Sun, Guolei and Zhang, Kai and Van Gool, Luc and Timofte, Radu},
  booktitle={Proceedings of the IEEE/CVF international conference on computer vision},
  pages={1833--1844},
  year={2021}
}

@inproceedings{chen2021pre,
  title={Pre-trained image processing transformer},
  author={Chen, Hanting and Wang, Yunhe and Guo, Tianyu and Xu, Chang and Deng, Yiping and Liu, Zhenhua and Ma, Siwei and Xu, Chunjing and Xu, Chao and Gao, Wen},
  booktitle={Proceedings of the IEEE/CVF conference on computer vision and pattern recognition},
  pages={12299--12310},
  year={2021}
}

@inproceedings{chen2023activating,
  title={Activating more pixels in image super-resolution transformer},
  author={Chen, Xiangyu and Wang, Xintao and Zhou, Jiantao and Qiao, Yu and Dong, Chao},
  booktitle={Proceedings of the IEEE/CVF conference on computer vision and pattern recognition},
  pages={22367--22377},
  year={2023}
}

@inproceedings{zhang2022efficient,
  title={Efficient long-range attention network for image super-resolution},
  author={Zhang, Xindong and Zeng, Hui and Guo, Shi and Zhang, Lei},
  booktitle={European conference on computer vision},
  pages={649--667},
  year={2022},
  organization={Springer}
}

@inproceedings{wang2021real,
  title={Real-esrgan: Training real-world blind super-resolution with pure synthetic data},
  author={Wang, Xintao and Xie, Liangbin and Dong, Chao and Shan, Ying},
  booktitle={Proceedings of the IEEE/CVF international conference on computer vision},
  pages={1905--1914},
  year={2021}
}

@inproceedings{zhang2021designing,
  title={Designing a practical degradation model for deep blind image super-resolution},
  author={Zhang, Kai and Liang, Jingyun and Van Gool, Luc and Timofte, Radu},
  booktitle={Proceedings of the IEEE/CVF international conference on computer vision},
  pages={4791--4800},
  year={2021}
}

@inproceedings{dai2019second,
  title={Second-order attention network for single image super-resolution},
  author={Dai, Tao and Cai, Jianrui and Zhang, Yongbing and Xia, Shu-Tao and Zhang, Lei},
  booktitle={Proceedings of the IEEE/CVF conference on computer vision and pattern recognition},
  pages={11065--11074},
  year={2019}
}

@inproceedings{wang2018esrgan,
  title={Esrgan: Enhanced super-resolution generative adversarial networks},
  author={Wang, Xintao and Yu, Ke and Wu, Shixiang and Gu, Jinjin and Liu, Yihao and Dong, Chao and Qiao, Yu and Change Loy, Chen},
  booktitle={Proceedings of the European conference on computer vision (ECCV) workshops},
  pages={0--0},
  year={2018}
}

@article{wang2024exploiting,
  title={Exploiting diffusion prior for real-world image super-resolution},
  author={Wang, Jianyi and Yue, Zongsheng and Zhou, Shangchen and Chan, Kelvin CK and Loy, Chen Change},
  journal={International Journal of Computer Vision},
  volume={132},
  number={12},
  pages={5929--5949},
  year={2024},
  publisher={Springer}
}

@inproceedings{lin2024diffbir,
  title={Diffbir: Toward blind image restoration with generative diffusion prior},
  author={Lin, Xinqi and He, Jingwen and Chen, Ziyan and Lyu, Zhaoyang and Dai, Bo and Yu, Fanghua and Qiao, Yu and Ouyang, Wanli and Dong, Chao},
  booktitle={European conference on computer vision},
  pages={430--448},
  year={2024},
  organization={Springer}
}

@inproceedings{yu2024scaling,
  title={Scaling up to excellence: Practicing model scaling for photo-realistic image restoration in the wild},
  author={Yu, Fanghua and Gu, Jinjin and Li, Zheyuan and Hu, Jinfan and Kong, Xiangtao and Wang, Xintao and He, Jingwen and Qiao, Yu and Dong, Chao},
  booktitle={Proceedings of the IEEE/CVF conference on computer vision and pattern recognition},
  pages={25669--25680},
  year={2024}
}

@inproceedings{yang2024pixel,
  title={Pixel-aware stable diffusion for realistic image super-resolution and personalized stylization},
  author={Yang, Tao and Wu, Rongyuan and Ren, Peiran and Xie, Xuansong and Zhang, Lei},
  booktitle={European conference on computer vision},
  pages={74--91},
  year={2024},
  organization={Springer}
}

@inproceedings{wu2024seesr,
  title={Seesr: Towards semantics-aware real-world image super-resolution},
  author={Wu, Rongyuan and Yang, Tao and Sun, Lingchen and Zhang, Zhengqiang and Li, Shuai and Zhang, Lei},
  booktitle={Proceedings of the IEEE/CVF conference on computer vision and pattern recognition},
  pages={25456--25467},
  year={2024}
}

@article{ai2024dreamclear,
  title={DreamClear: High-Capacity Real-World Image Restoration with Privacy-Safe Dataset Curation},
  author={Ai, Yuang and Zhou, Xiaoqiang and Huang, Huaibo and Han, Xiaotian and Chen, Zhengyu and You, Quanzeng and Yang, Hongxia},
  journal={Advances in Neural Information Processing Systems},
  volume={37},
  pages={55443--55469},
  year={2024}
}

@inproceedings{chen2025faithdiff,
  title={Faithdiff: Unleashing diffusion priors for faithful image super-resolution},
  author={Chen, Junyang and Pan, Jinshan and Dong, Jiangxin},
  booktitle={Proceedings of the Computer Vision and Pattern Recognition Conference},
  pages={28188--28197},
  year={2025}
}

@inproceedings{duan2025dit4sr,
  title={Dit4sr: Taming diffusion transformer for real-world image super-resolution},
  author={Duan, Zheng-Peng and Zhang, Jiawei and Jin, Xin and Zhang, Ziheng and Xiong, Zheng and Zou, Dongqing and Ren, Jimmy S and Guo, Chunle and Li, Chongyi},
  booktitle={Proceedings of the IEEE/CVF International Conference on Computer Vision},
  pages={18948--18958},
  year={2025}
}

@article{wu2024one,
  title={One-step effective diffusion network for real-world image super-resolution},
  author={Wu, Rongyuan and Sun, Lingchen and Ma, Zhiyuan and Zhang, Lei},
  journal={Advances in Neural Information Processing Systems},
  volume={37},
  pages={92529--92553},
  year={2024}
}

@inproceedings{cai2019toward,
  title={Toward real-world single image super-resolution: A new benchmark and a new model},
  author={Cai, Jianrui and Zeng, Hui and Yong, Hongwei and Cao, Zisheng and Zhang, Lei},
  booktitle={Proceedings of the IEEE/CVF international conference on computer vision},
  pages={3086--3095},
  year={2019}
}

@inproceedings{wei2020component,
  title={Component divide-and-conquer for real-world image super-resolution},
  author={Wei, Pengxu and Xie, Ziwei and Lu, Hannan and Zhan, Zongyuan and Ye, Qixiang and Zuo, Wangmeng and Lin, Liang},
  booktitle={European conference on computer vision},
  pages={101--117},
  year={2020},
  organization={Springer}
}

@inproceedings{zhang2018unreasonable,
  title={The unreasonable effectiveness of deep features as a perceptual metric},
  author={Zhang, Richard and Isola, Phillip and Efros, Alexei A and Shechtman, Eli and Wang, Oliver},
  booktitle={Proceedings of the IEEE conference on computer vision and pattern recognition},
  pages={586--595},
  year={2018}
}

@inproceedings{wang2023exploring,
  title={Exploring clip for assessing the look and feel of images},
  author={Wang, Jianyi and Chan, Kelvin CK and Loy, Chen Change},
  booktitle={Proceedings of the AAAI conference on artificial intelligence},
  volume={37},
  number={2},
  pages={2555--2563},
  year={2023}
}

@inproceedings{yang2022maniqa,
  title={Maniqa: Multi-dimension attention network for no-reference image quality assessment},
  author={Yang, Sidi and Wu, Tianhe and Shi, Shuwei and Lao, Shanshan and Gong, Yuan and Cao, Mingdeng and Wang, Jiahao and Yang, Yujiu},
  booktitle={Proceedings of the IEEE/CVF conference on computer vision and pattern recognition},
  pages={1191--1200},
  year={2022}
}

@inproceedings{ke2021musiq,
  title={Musiq: Multi-scale image quality transformer},
  author={Ke, Junjie and Wang, Qifei and Wang, Yilin and Milanfar, Peyman and Yang, Feng},
  booktitle={Proceedings of the IEEE/CVF international conference on computer vision},
  pages={5148--5157},
  year={2021}
}

@article{wang2004image,
  title={Image quality assessment: from error visibility to structural similarity},
  author={Wang, Zhou and Bovik, Alan C and Sheikh, Hamid R and Simoncelli, Eero P},
  journal={IEEE transactions on image processing},
  volume={13},
  number={4},
  pages={600--612},
  year={2004},
  publisher={IEEE}
}

@misc{flux2024,
  author = {{Black Forest Labs}},
  title = {Flux},
  howpublished = {\url{https://github.com/black-forest-labs/flux}},
  year = {2024},
  note = {Accessed: 2024}
}

@inproceedings{blau2018perception,
  title={The perception-distortion tradeoff},
  author={Blau, Yochai and Michaeli, Tomer},
  booktitle={Proceedings of the IEEE conference on computer vision and pattern recognition},
  pages={6228--6237},
  year={2018}
}

@inproceedings{jinjin2020pipal,
  title={Pipal: a large-scale image quality assessment dataset for perceptual image restoration},
  author={Jinjin, Gu and Haoming, Cai and Haoyu, Chen and Xiaoxing, Ye and Ren, Jimmy S and Chao, Dong},
  booktitle={European conference on computer vision},
  pages={633--651},
  year={2020},
  organization={Springer}
}

@article{hu2022lora,
  title={Lora: Low-rank adaptation of large language models.},
  author={Hu, Edward J and Shen, Yelong and Wallis, Phillip and Allen-Zhu, Zeyuan and Li, Yuanzhi and Wang, Shean and Wang, Lu and Chen, Weizhu and others},
  journal={ICLR},
  volume={1},
  number={2},
  pages={3},
  year={2022}
}

@article{ho2020denoising,
  title={Denoising diffusion probabilistic models},
  author={Ho, Jonathan and Jain, Ajay and Abbeel, Pieter},
  journal={Advances in neural information processing systems},
  volume={33},
  pages={6840--6851},
  year={2020}
}

@inproceedings{rombach2022high,
  title={High-resolution image synthesis with latent diffusion models},
  author={Rombach, Robin and Blattmann, Andreas and Lorenz, Dominik and Esser, Patrick and Ommer, Bj{\"o}rn},
  booktitle={Proceedings of the IEEE/CVF conference on computer vision and pattern recognition},
  pages={10684--10695},
  year={2022}
}

@inproceedings{esser2024scaling,
  title={Scaling rectified flow transformers for high-resolution image synthesis},
  author={Esser, Patrick and Kulal, Sumith and Blattmann, Andreas and Entezari, Rahim and M{\"u}ller, Jonas and Saini, Harry and Levi, Yam and Lorenz, Dominik and Sauer, Axel and Boesel, Frederic and others},
  booktitle={Forty-first international conference on machine learning},
  year={2024}
}

@article{ramesh2022hierarchical,
  title={Hierarchical text-conditional image generation with clip latents},
  author={Ramesh, Aditya and Dhariwal, Prafulla and Nichol, Alex and Chu, Casey and Chen, Mark},
  journal={arXiv preprint arXiv:2204.06125},
  volume={1},
  number={2},
  pages={3},
  year={2022}
}

@article{dhariwal2021diffusion,
  title={Diffusion models beat gans on image synthesis},
  author={Dhariwal, Prafulla and Nichol, Alexander},
  journal={Advances in neural information processing systems},
  volume={34},
  pages={8780--8794},
  year={2021}
}

@article{podell2023sdxl,
  title={Sdxl: Improving latent diffusion models for high-resolution image synthesis},
  author={Podell, Dustin and English, Zion and Lacey, Kyle and Blattmann, Andreas and Dockhorn, Tim and M{\"u}ller, Jonas and Penna, Joe and Rombach, Robin},
  journal={arXiv preprint arXiv:2307.01952},
  year={2023}
}

@article{bar2023multidiffusion,
  title={Multidiffusion: Fusing diffusion paths for controlled image generation},
  author={Bar-Tal, Omer and Yariv, Lior and Lipman, Yaron and Dekel, Tali},
  year={2023}
}

@inproceedings{chan2021glean,
  title={Glean: Generative latent bank for large-factor image super-resolution},
  author={Chan, Kelvin CK and Wang, Xintao and Xu, Xiangyu and Gu, Jinwei and Loy, Chen Change},
  booktitle={Proceedings of the IEEE/CVF conference on computer vision and pattern recognition},
  pages={14245--14254},
  year={2021}
}

@inproceedings{wang2021towards,
  title={Towards real-world blind face restoration with generative facial prior},
  author={Wang, Xintao and Li, Yu and Zhang, Honglun and Shan, Ying},
  booktitle={Proceedings of the IEEE/CVF conference on computer vision and pattern recognition},
  pages={9168--9178},
  year={2021}
}

@inproceedings{ledig2017photo,
  title={Photo-realistic single image super-resolution using a generative adversarial network},
  author={Ledig, Christian and Theis, Lucas and Husz{\'a}r, Ferenc and Caballero, Jose and Cunningham, Andrew and Acosta, Alejandro and Aitken, Andrew and Tejani, Alykhan and Totz, Johannes and Wang, Zehan and others},
  booktitle={Proceedings of the IEEE conference on computer vision and pattern recognition},
  pages={4681--4690},
  year={2017}
}

@inproceedings{deng2025acquire,
  title={Acquire and then Adapt: Squeezing out Text-to-Image Model for Image Restoration},
  author={Deng, Junyuan and Wu, Xinyi and Yang, Yongxing and Zhu, Congchao and Wang, Song and Wu, Zhenyao},
  booktitle={Proceedings of the Computer Vision and Pattern Recognition Conference},
  pages={23195--23206},
  year={2025}
}

@article{yue2024efficient,
  title={Efficient diffusion model for image restoration by residual shifting},
  author={Yue, Zongsheng and Wang, Jianyi and Loy, Chen Change},
  journal={IEEE Transactions on Pattern Analysis and Machine Intelligence},
  year={2024},
  publisher={IEEE}
}

@inproceedings{dong2025tsd,
  title={Tsd-sr: One-step diffusion with target score distillation for real-world image super-resolution},
  author={Dong, Linwei and Fan, Qingnan and Guo, Yihong and Wang, Zhonghao and Zhang, Qi and Chen, Jinwei and Luo, Yawei and Zou, Changqing},
  booktitle={Proceedings of the Computer Vision and Pattern Recognition Conference},
  pages={23174--23184},
  year={2025}
}

@article{liu2023visual,
  title={Visual instruction tuning},
  author={Liu, Haotian and Li, Chunyuan and Wu, Qingyang and Lee, Yong Jae},
  journal={Advances in neural information processing systems},
  volume={36},
  pages={34892--34916},
  year={2023}
}

@article{chen2023pixart,
  title={Pixart-$\alpha$: Fast training of diffusion transformer for photorealistic text-to-image synthesis},
  author={Chen, Junsong and Yu, Jincheng and Ge, Chongjian and Yao, Lewei and Xie, Enze and Wu, Yue and Wang, Zhongdao and Kwok, James and Luo, Ping and Lu, Huchuan and others},
  journal={arXiv preprint arXiv:2310.00426},
  year={2023}
}

@article{saharia2022photorealistic,
  title={Photorealistic text-to-image diffusion models with deep language understanding},
  author={Saharia, Chitwan and Chan, William and Saxena, Saurabh and Li, Lala and Whang, Jay and Denton, Emily L and Ghasemipour, Kamyar and Gontijo Lopes, Raphael and Karagol Ayan, Burcu and Salimans, Tim and others},
  journal={Advances in neural information processing systems},
  volume={35},
  pages={36479--36494},
  year={2022}
}

@article{xie2024sana,
  title={Sana: Efficient high-resolution image synthesis with linear diffusion transformers},
  author={Xie, Enze and Chen, Junsong and Chen, Junyu and Cai, Han and Tang, Haotian and Lin, Yujun and Zhang, Zhekai and Li, Muyang and Zhu, Ligeng and Lu, Yao and others},
  journal={arXiv preprint arXiv:2410.10629},
  year={2024}
}

@inproceedings{liang2022details,
  title={Details or artifacts: A locally discriminative learning approach to realistic image super-resolution},
  author={Liang, Jie and Zeng, Hui and Zhang, Lei},
  booktitle={Proceedings of the IEEE/CVF conference on computer vision and pattern recognition},
  pages={5657--5666},
  year={2022}
}

@inproceedings{liang2022efficient,
  title={Efficient and degradation-adaptive network for real-world image super-resolution},
  author={Liang, Jie and Zeng, Hui and Zhang, Lei},
  booktitle={European Conference on Computer Vision},
  pages={574--591},
  year={2022},
  organization={Springer}
}

@article{xie2023desra,
  title={Desra: detect and delete the artifacts of gan-based real-world super-resolution models},
  author={Xie, Liangbin and Wang, Xintao and Chen, Xiangyu and Li, Gen and Shan, Ying and Zhou, Jiantao and Dong, Chao},
  journal={arXiv preprint arXiv:2307.02457},
  year={2023}
}

@article{yue2023resshift,
  title={Resshift: Efficient diffusion model for image super-resolution by residual shifting},
  author={Yue, Zongsheng and Wang, Jianyi and Loy, Chen Change},
  journal={Advances in Neural Information Processing Systems},
  volume={36},
  pages={13294--13307},
  year={2023}
}

@article{lin2025harnessing,
  title={Harnessing diffusion-yielded score priors for image restoration},
  author={Lin, Xinqi and Yu, Fanghua and Hu, Jinfan and You, Zhiyuan and Shi, Wu and Ren, Jimmy S and Gu, Jinjin and Dong, Chao},
  journal={arXiv preprint arXiv:2507.20590},
  year={2025}
}

@inproceedings{chen2025adversarial,
  title={Adversarial diffusion compression for real-world image super-resolution},
  author={Chen, Bin and Li, Gehui and Wu, Rongyuan and Zhang, Xindong and Chen, Jie and Zhang, Jian and Zhang, Lei},
  booktitle={Proceedings of the Computer Vision and Pattern Recognition Conference},
  pages={28208--28220},
  year={2025}
}

@inproceedings{chen2022real,
  title={Real-world blind super-resolution via feature matching with implicit high-resolution priors},
  author={Chen, Chaofeng and Shi, Xinyu and Qin, Yipeng and Li, Xiaoming and Han, Xiaoguang and Yang, Tao and Guo, Shihui},
  booktitle={Proceedings of the 30th ACM International Conference on Multimedia},
  pages={1329--1338},
  year={2022}
}

@inproceedings{chen2023dual,
  title={Dual aggregation transformer for image super-resolution},
  author={Chen, Zheng and Zhang, Yulun and Gu, Jinjin and Kong, Linghe and Yang, Xiaokang and Yu, Fisher},
  booktitle={Proceedings of the IEEE/CVF international conference on computer vision},
  pages={12312--12321},
  year={2023}
}

@inproceedings{cheng2025effective,
  title={Effective diffusion transformer architecture for image super-resolution},
  author={Cheng, Kun and Yu, Lei and Tu, Zhijun and He, Xiao and Chen, Liyu and Guo, Yong and Zhu, Mingrui and Wang, Nannan and Gao, Xinbo and Hu, Jie},
  booktitle={Proceedings of the AAAI Conference on Artificial Intelligence},
  volume={39},
  number={3},
  pages={2455--2463},
  year={2025}
}

@article{cui2024taming,
  title={Taming diffusion prior for image super-resolution with domain shift sdes},
  author={Cui, Qinpeng and Liu, Yixuan and Zhang, Xinyi and Bao, Qiqi and Liao, Qingmin and Wang, Li and Lu, Tian and Liu, Zicheng and Wang, Zhongdao and Barsoum, Emad},
  journal={arXiv preprint arXiv:2409.17778},
  year={2024}
}

@article{gu2024consissr,
  title={Consissr: Delving deep into consistency in diffusion-based image super-resolution},
  author={Gu, Junhao and Jiang, Peng-Tao and Zhang, Hao and Zhou, Mi and Chen, Jinwei and Yang, Wenming and Li, Bo},
  year={2024}
}

@article{li2024distillation,
  title={Distillation-free one-step diffusion for real-world image super-resolution},
  author={Li, Jianze and Cao, Jiezhang and Zou, Zichen and Su, Xiongfei and Yuan, Xin and Zhang, Yulun and Guo, Yong and Yang, Xiaokang},
  year={2024}
}

@article{li2025one,
  title={One diffusion step to real-world super-resolution via flow trajectory distillation},
  author={Li, Jianze and Cao, Jiezhang and Guo, Yong and Li, Wenbo and Zhang, Yulun},
  journal={arXiv preprint arXiv:2502.01993},
  year={2025}
}

@inproceedings{noroozi2024you,
  title={You only need one step: Fast super-resolution with stable diffusion via scale distillation},
  author={Noroozi, Mehdi and Hadji, Isma and Martinez, Brais and Bulat, Adrian and Tzimiropoulos, Georgios},
  booktitle={European Conference on Computer Vision},
  pages={145--161},
  year={2024},
  organization={Springer}
}

@inproceedings{qu2024xpsr,
  title={Xpsr: Cross-modal priors for diffusion-based image super-resolution},
  author={Qu, Yunpeng and Yuan, Kun and Zhao, Kai and Xie, Qizhi and Hao, Jinhua and Sun, Ming and Zhou, Chao},
  booktitle={European Conference on Computer Vision},
  pages={285--303},
  year={2024},
  organization={Springer}
}

@inproceedings{sun2024coser,
  title={Coser: Bridging image and language for cognitive super-resolution},
  author={Sun, Haoze and Li, Wenbo and Liu, Jianzhuang and Chen, Haoyu and Pei, Renjing and Zou, Xueyi and Yan, Youliang and Yang, Yujiu},
  booktitle={Proceedings of the IEEE/CVF Conference on Computer Vision and Pattern Recognition},
  pages={25868--25878},
  year={2024}
}

@inproceedings{sun2025pixel,
  title={Pixel-level and semantic-level adjustable super-resolution: A dual-lora approach},
  author={Sun, Lingchen and Wu, Rongyuan and Ma, Zhiyuan and Liu, Shuaizheng and Yi, Qiaosi and Zhang, Lei},
  booktitle={Proceedings of the Computer Vision and Pattern Recognition Conference},
  pages={2333--2343},
  year={2025}
}

@article{sun2024improving,
  title={Improving the stability of diffusion models for content consistent super-resolution},
  author={Sun, Lingchen and Wu, Rongyuan and Zhang, Zhengqiang and Yong, Hongwei and Zhang, Lei},
  journal={CoRR},
  year={2024}
}

@article{tsao2024holisdip,
  title={Holisdip: Image super-resolution via holistic semantics and diffusion prior},
  author={Tsao, Li-Yuan and Chen, Hao-Wei and Chung, Hao-Wei and Sun, Deqing and Lee, Chun-Yi and Chan, Kelvin CK and Yang, Ming-Hsuan},
  journal={arXiv preprint arXiv:2411.18662},
  year={2024}
}

@article{wan2024clearsr,
  title={Clearsr: Latent low-resolution image embeddings help diffusion-based real-world super resolution models see clearer},
  author={Wan, Yuhao and Jiang, Peng-Tao and Hou, Qibin and Zhang, Hao and Chen, Jinwei and Cheng, Ming-Ming and Li, Bo},
  year={2024}
}

@inproceedings{wang2024sinsr,
  title={Sinsr: diffusion-based image super-resolution in a single step},
  author={Wang, Yufei and Yang, Wenhan and Chen, Xinyuan and Wang, Yaohui and Guo, Lanqing and Chau, Lap-Pui and Liu, Ziwei and Qiao, Yu and Kot, Alex C and Wen, Bihan},
  booktitle={Proceedings of the IEEE/CVF conference on computer vision and pattern recognition},
  pages={25796--25805},
  year={2024}
}

@article{xie2024addsr,
  title={Addsr: Accelerating diffusion-based blind super-resolution with adversarial diffusion distillation},
  author={Xie, Rui and Zhao, Chen and Zhang, Kai and Zhang, Zhenyu and Zhou, Jun and Yang, Jian and Tai, Ying},
  journal={arXiv preprint arXiv:2404.01717},
  year={2024}
}

@article{zhang2024degradation,
  title={Degradation-guided one-step image super-resolution with diffusion priors},
  author={Zhang, Aiping and Yue, Zongsheng and Pei, Renjing and Ren, Wenqi and Cao, Xiaochun},
  journal={arXiv preprint arXiv:2409.17058},
  year={2024}
}

@inproceedings{li2023lsdir,
  title={Lsdir: A large scale dataset for image restoration},
  author={Li, Yawei and Zhang, Kai and Liang, Jingyun and Cao, Jiezhang and Liu, Ce and Gong, Rui and Zhang, Yulun and Tang, Hao and Liu, Yun and Demandolx, Denis and others},
  booktitle={Proceedings of the IEEE/CVF Conference on Computer Vision and Pattern Recognition},
  pages={1775--1787},
  year={2023}
}

@inproceedings{karras2019style,
  title={A style-based generator architecture for generative adversarial networks},
  author={Karras, Tero and Laine, Samuli and Aila, Timo},
  booktitle={Proceedings of the IEEE/CVF conference on computer vision and pattern recognition},
  pages={4401--4410},
  year={2019}
}

@inproceedings{liu2025patchscaler,
  title={PatchScaler: An Efficient Patch-Independent Diffusion Model for Image Super-Resolution},
  author={Liu, Yong and Dong, Hang and Pan, Jinshan and Dong, Qingji and Chen, Kai and Zhang, Rongxiang and Fu, Lean and Wang, Fei},
  booktitle={Proceedings of the IEEE/CVF International Conference on Computer Vision},
  pages={11283--11293},
  year={2025}
}

@inproceedings{liu2023unfolding,
  title={Unfolding once is enough: A deployment-friendly transformer unit for super-resolution},
  author={Liu, Yong and Dong, Hang and Liang, Boyang and Liu, Songwei and Dong, Qingji and Chen, Kai and Chen, Fangmin and Fu, Lean and Wang, Fei},
  booktitle={Proceedings of the 31st ACM international conference on multimedia},
  pages={7952--7960},
  year={2023}
}
}

\end{document}